\documentclass{article}
\usepackage{ijcai17}

\usepackage{times}
\usepackage{times}
\usepackage{algorithm}
\usepackage{algorithmic}
\usepackage{xcolor}
\usepackage[cmex10]{amsmath}
\usepackage{graphicx}
\usepackage{multirow}
\usepackage{amsfonts}
\usepackage{multirow}
\usepackage{booktabs}
\usepackage{subfigure}

\usepackage{amsmath,amsthm}

\DeclareMathOperator*{\argmin}{argmin}

\renewcommand{\algorithmicrequire}{ \textbf{Input:}}      %Use Input in the format of Algorithm
\renewcommand{\algorithmicensure}{ \textbf{Output:}}     %UseOutput in the format of Algorithm
\title{Towards well-specified semi-supervised model-based classifiers via structural adaptation}%\thanks{These match the formatting instructions of IJCAI-07. The support of IJCAI, Inc. is acknowledged.}}
\author{Zhaocai Sun, William K. Cheung, Xiaofeng Zhang\\
Harbin Institute of Technology Shenzhen Graduate School, Shenzhen, China\\
Department of Computer Science, Hong Kong Baptist University, Hong Kong  \\
\{69180577@qq.com,william@comp.hkbu.edu.hk,zhangxiaofeng@hit.edu.cn\}
}

\begin{document}

\maketitle

\begin{abstract}

Semi-supervised learning plays an important role in large-scale machine learning. Properly using additional unlabeled data (largely available nowadays) often can improve the machine learning accuracy. However, if the machine learning model is misspecified for the underlying true data distribution, the model performance could be seriously jeopardized. This issue is known as model misspecification. To address this issue, we focus on generative models and propose a criterion to detect the onset of model misspecification by measuring the performance difference between models obtained using supervised and semi-supervised learning. Then, we propose to automatically modify the generative models during model training to achieve an unbiased generative model. Rigorous experiments were carried out to evaluate the proposed method using two image classification data sets PASCAL VOC'07 and MIR Flickr. Our proposed method has been demonstrated to outperform a number of state-of-the-art semi-supervised learning approaches for the classification task.
%The comparison results on sophisticated image data (the PASCAL
%VOC��07 set and the MIR Flickr set) show our modified model enormously superior to other generative semi-supervised models, and even outperforms most the-date-of-art non-generative semi-supervised models.
\end{abstract}

\section{Introduction}

Semi-supervised learning (SSL) plays an important role in many real world machine learning applications, such as image classification \cite{Papandreou2015Weakly}, speech recognition \cite{Cui2012Multi}, and text categorization \cite{liu2016semi,Sun2012Batch}, where the cost of annotating unlabeled data is generally considered too high to afford. SSL tries to make use of additional unlabeled data together with a limited amount of labeled data to enhance model learning accuracy \cite{Zhu:ICML:03}, and thus has gained a lot of researchers' attention \cite{Priebe:TPAMI:2011,Fox-Roberts:2014:JMLR}.

However, it is generally believed that using additional unlabeled data for SSL does not always guarantee increase in model learning accuracy. Sometimes, it may end up with performance degradation.
%sometimes which contradicts aforementioned expectation, i.e.,  if more unlabeled data are used, then the model performance of SSL should increase \cite{Priebe:TPAMI:2011}.
This phenomenon is known as model misspecification \cite{Priebe:TPAMI:2011,Fox-Roberts:2014:JMLR,Bach:2006:NIPS} or safe semi-supervised learning \cite{Zhou:KIS:2010}.
%, and which already becomes an emerging hot research issue.
Yet in essence, model misspecification and safe semi-supervised learning are two concepts proposed independently, despite their common goal of designing SSL approach so that its performance, even in the worst case, is still better than that of the simple supervised learning approach \cite{Zhou:AAAI16,Zhou:AAAI17}. For safe semi-supervised learning, most of the existing approaches were proposed for non-parametric models \cite{Zhou:AAAI16,Li2011Towards}. For instance, in \cite{Zhou:AAAI16},
%this issue, the existing related works can be classified mainly into two lines, i.e., generative model based approach and non-parametric model based approach. For non-parametric SSL approaches such as \cite{}, they
they first set a baseline classifier corresponding to the supervised learners in the worst cases, and then further optimize the performance of the classifier using the proposed SSL method. The performance difference between the classifiers learned in supervised and semi-supervised manners is crucial in designing such SSL algorithms.

Approaches proposed from the perspective of model misspecification are mostly for parametric models \cite{Priebe:TPAMI:2011,loog2015semi,loog2016contrastive} where the mismatch between the unlabeled data and the employed generative model are often used to guide the model learning. For instance,  the mixture model is adopted in \cite{Priebe:TPAMI:2011} to represent both labeled and unlabeled data, which then formulates the corresponding Bayes plug-in classifier based on the mixture density functions. The performance of this Bayes plug-in classifier seriously relies on how far the learnt parameters are from the true ones. The performance degradation mainly comes from the model bias and estimation error. While how the unlabeled data could affect the SSL model performance is theoretically analysed in \cite{Priebe:TPAMI:2011}, how to detect the onset of the model misspecification and how to solve the challenge have not been addressed yet. Along this line, the lower bound and upper bound of the performance of semi-supervised generative models are analyzed in \cite{Fox-Roberts:2014:JMLR}, and then the authors propose to use the ratio of unlabeled data to labeled data to control the model performance. Inspired by these two works, we propose a generative model based approach for addressing the model misspecification issue. Different from the two aforementioned methods, we not only explore how to detect whether the model misspecification occurs, but also propose a model modification method instead of controlling the unlabeled data to be utilized.

The remaining of this paper is organized as follows. We first present related work in Section \ref{sec:Related} and formulate the model misspecification problem using generative models in Section \ref{sec:Problem}. The proposed SSL learning approach is described in Section \ref{sec:Approach}. Experimental results based on two image classification datasets are reported in Section \ref{sec:Exp}. We conclude the paper in Section \ref{sec:Conclusion}.

\section{Related Work}\label{sec:Related}

The difficulty to achieve a reliable semi-supervised learning method has been reported in some earlier works (e.g., \cite{cozman2002unlabeled}). The general believe is that a better SSL model is not always guaranteed even though additional unlabeled data are used for the model learning. There exist a number of factors that may affect the SSL performance such as the quality of the training data as well as the classifier itself \cite{Priebe:TPAMI:2011}. Some researchers considered that to be the consequence of a wrong model assumption \cite{Wang:TNNLS:12}. Practically, it is hard to assume a perfect generative model without any prior knowledge about the unknown data set. In \cite{Zhou:KIS:2010,Zhou:AAAI17}, the underlying challenge is viewed as having some unlabeled data assigned with incorrect labels which are then used to augment the labeled training data set.

Some recent research works proposed safety-aware mechanisms to restrict the SSL from using risky unlabeled data \cite{safety:TNN:15,Zhou:AAAI17,Zhou:AAAI16}. Other than the earlier disagreement based SSL \cite{bennett1998semi}, several safe SSL approaches have been proposed, such as S3VMs \cite{bennett1998semi}, S4VM \cite{Li2011Towards} and UMVP \cite{Zhou:AAAI16} with promising experimental results. However, most of them fall into the non-parametric category. To leverage on their good generalization capabilities, generative models based on SSL are common for many applications such as image classification \cite{Dao:Neuro:2017}. The two most representative works addressing the model misspecification issue include \cite{Priebe:TPAMI:2011} and \cite{Fox-Roberts:2014:JMLR}.
In \cite{Priebe:TPAMI:2011},
%is the milestone paper which gives some theoretical analysis about how unlabeled data could affect SSL performance.
%In this work,
they started with the asymptotic optimal parameters of two generative models obtained using fully supervised learning and fully unsupervised learning, respectively. Then, they proved that if the KL divergence between the distributions generated by the two generative models is small, the SSL performance is less likely to be affected by the addition of unlabeled data. More theoretical analysis was provided in \cite{Fox-Roberts:2014:JMLR} where the local and global bounds on the divergence of SSL are formulated. Then, they proposed an unbiased generative SSL which is defined based on an unbiased likelihood estimator exponentially controlled by the ratio
%$N_l/N$
of the number of labeled data over the total number of data.

Our work is different from \cite{Priebe:TPAMI:2011} and \cite{Fox-Roberts:2014:JMLR} as follows. We focus more on adaptively modifying the structure of the generative models instead of controlling risky data to be used. Furthermore, we propose a criterion to determine whether a model misspecification occurs or not. To the best of our knowledge, neither approaches have been considered in the literature.

\section{Model Misspecification Problem}\label{sec:Problem}

\begin{figure}[!htb]%\label{fig:model}
\centering
\subfigure[When model is misspecified.]{
%\label{fig.yalebwrong.a}
\includegraphics[width=2.4in]{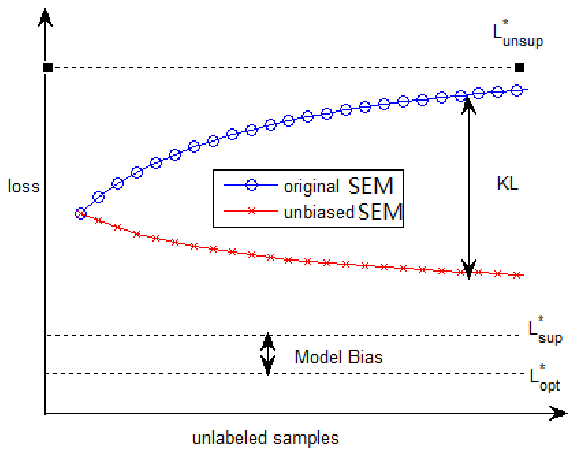}}
\subfigure[When model is well-specified.]{
%\label{fig.yalebwrong.b}
\includegraphics[width=2.4in]{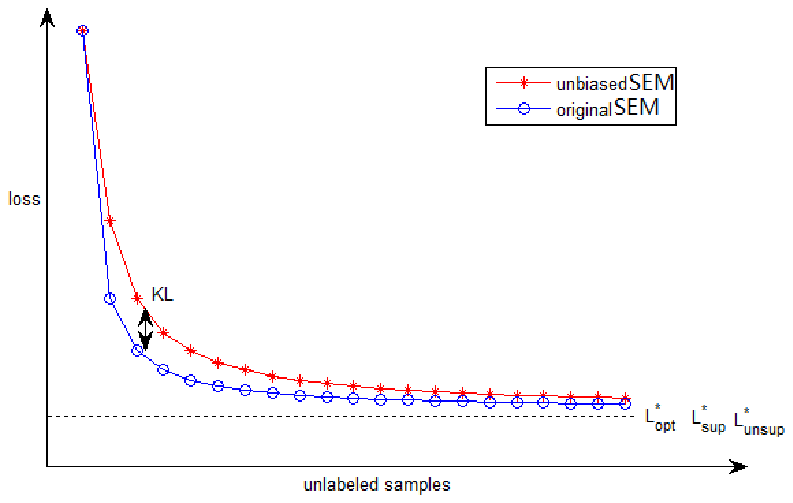}}
\caption{The illustrating example of the proposed approach.}
\label{fig:model}
\end{figure}

Given a finite number of labeled data and an infinite number of unlabeled data, it is impossible to directly detect whether the model misspecification  occurs or not as the true data distribution is generally unknown. Figure \ref{fig:model} gives an illustrating example of the proposed approach. Assume that the model estimation error is ignored. In Figure \ref{fig:model} (a), the assumed semi-supervised generative model (SEM) is misspecified, and its performance, in the worst case, would converge to the classification loss bound $L_{unsup}^*$ of SEM learnt in an unsupervised manner if infinite unlabeled data were used. And the unbiased SEM \cite{Fox-Roberts:2014:JMLR} would converge to the classification loss bound $L_{sup}^*$ of SEM learnt in the supervised manner. However, these classification loss bounds are higher than the best classification loss $L_{opt}^{*}$ obtained by a well-specified model. Therefore, the minimum model bias from the learnt SEMs to the true data model can be approximated by $L_{opt}^{*} - L_{sup}^*$ which is indicated by the loss difference, i.e., $L_{sup}^* - L_{unsup}^*$. Such loss difference can be approximated by the KL distance between two SEMs. If the assumed SEM is well-specified as plotted in Figure \ref{fig:model} (b), the classification loss bound of the SEM, unbiased SEM and the well-specified SEM will be the same, i.e., $L_{opt}^{*} = L_{sup}^* = L_{unsup}^*$, in the ideal case. Consequently, the model difference (KL) of two SEMs should be also small. Inspired by this observation, we conjecture that there must exist the model misspecification if the KL between two SEMs is getting large and therefore we can use this value to determine whether a model misspecification occurs or not. The problem is formulated in the following paragraphs.
%In this case, the classification loss $L_{sup}^{*}$ of supervised learning is still higher than the loss, $L_{opt}^{*}$, of a well-specified model, as illustrated in Figure \ref{fig:model} (a). If the generative model is well-specified, the ideal estimations of unsupervised generative model and supervised generative model will be the same. Therefore, the original SEM and the unbiased SEM will converge to the same value. This value is also the best estimation of the well-specified model, as illustrated in Figure \ref{fig:model} (b). Therefore from Figure \ref{fig:model} (a), if the difference of classification loss, i.e., $L_{sup}^{*}-L_{unsup}^{*}$, is getting large, we can conjecture there must exist the model misspecification. Consequently, we can increase the model complexity by increasing the number of components. The modified models then gradually approximate to the true data models.
%\vspace{-0.4cm}
%\vspace{-0.4cm}

Let $X:=(x_1,x_2,\cdots, x_N)^T$ and $Y:=(y_1,y_2,\cdots,y_N)^T$ denote the training data and the corresponding label, respectively. The set of labeled data and unlabeled data are denoted as $S_l:=(X,Y)$ and $S_u :=(x^U_1,x^U_2,\cdots)^T$, respectively. The mixed set of labeled and unlabeled data is denoted as $D=S_l \bigcup S_u$. Suppose $P(X,Y)$ is the true data distribution for $D$ and $f(X,Y|\Theta)$ denote the assumed generative model with its parameter set as $\Theta$. A supervised generative model is obtained by learning the model parameters that best fit $P(X,Y)$, written as
%
%Supposing $P(X,Y)$ is the real but unknown distribution and $f(X,Y|\theta)$ is the assumed generative model with parameter $\theta$, the task of machine learning is to find out the best parameter  to simulate $P(X,Y)$. That is,

\begin{equation}
\label{eq:1}
 \min_{\Theta} KL(P(X,Y)\| f(X,Y|\Theta))
\end{equation}
where $KL(\cdot)$ is the Kullback-Leibler divergence, defined as
\begin{equation}\label{eq:2}\small
KL(P(X,Y)\| f(X,Y|\Theta)) = \int p(X,Y)\log\frac{p(X,Y)}{f(X,Y|\Theta)}dx
\end{equation}
%E���±��Բ��ԣ�
%-------------------
Given a generative model $f$ (e.g. GMM) with its parameter set $\Theta$,
a popular semi-supervised learning objective function is given as,

\begin{equation}\small
\label{eq:regsemi}
 \max \{\sum_{x_i\in S_l} \log (f(x_i,y_i|\Theta)) + \sum_{x_j\in S_u} \log (f(x_j|\Theta))\}
\end{equation}
%where $S_l$ and $S_u$ are the labeled and unlabeled training data set, respectively.

\newtheorem{thm}{Theorem}

\newtheorem{cor}{Corollary}
\begin{thm}
\label{theorem1}\small
If $|S_l| \rightarrow \infty$, $|S_u| \rightarrow \infty$, SSL problem in Eq.\ref{eq:regsemi} is equivalent to
\begin{equation}\label{eq:e4}
\small \min_{\Theta} \{KL(P(X,Y)\| f(X,Y|\Theta)) + \frac{|S_u|}{|S_l|}   KL(P(X,Y)\| f(X|\Theta))\}
\end{equation}
\end{thm}

\textbf{Proof.} For brevity, let $N_l = |S_l| $,$N_u = |S_u| $, $\lambda = N_u /N_l$, following the Theorem proposed in \cite{cozman2003semi}, the
MLE problem in Eq.\ref{eq:regsemi} is stated as:
\begin{equation}\small
\label{eq:eqv}
\max_{\Theta}\{ \frac{1}{1+\lambda}   E [\log f(X,Y|\Theta)] + \frac{\lambda}{1+\lambda} E [\log f(X|\Theta)]\}
\end{equation}
Using Eq.\ref{eq:2}, Eq.\ref{eq:eqv} is equivalent to Eq.\ref{eq:e4}.

\begin{cor}
If $\lambda\rightarrow \infty$, SSL problem as Eq. \ref{eq:e4} degenerates back to an unsupervised learning problem.
\end{cor}
%---------------
\begin{equation}\small
\label{eq:6}
 \min_{\Theta} KL(P(X,Y)\| f(X|\Theta))
\end{equation}
For this reason, \cite{Fox-Roberts:2014:JMLR} proposed a weighting strategy to avoid the model misspecification issue, given as
\begin{equation}\small
\label{eq:e7}
 \min_{\Theta} \{KL(P(X,Y)\| f(X,Y|\Theta)) +  KL(P(X,Y)\| f(X|\Theta))\}
\end{equation}
The effectiveness of this strategy is quite tricky as it seriously relies on whether there are enough labeled data or not. It becomes risky when the labeled data are few. Therefore, we propose a safer strategy as follows.
%However, \cite{Priebe:TPAMI:2011} shows that it is model misspecification causes the unsafe problem of semi-supervised learning.
%The work in \cite{} has not solved the essence of unsafe semi-supervised learning.

For a generative model $f$, let $\Theta_{sup}$ denote the parameter set learnt in a supervised manner using Eq. \ref{eq:1}. Similarly, $\Theta_{smsup}$, $\Theta_{usmsup}$, and $\Theta_{unsup}$ are the solutions to Eqs. \ref{eq:e4}, \ref{eq:e7}, and \ref{eq:6}, respectively, denoting original SSL models, unbiased SSL models, and unsupervised models. As the data set only contains a finite number of data points, the best estimations ${\Theta}^*_{sup}$,${\Theta}^*_{smsup}$,${\Theta}^*_{usmsup}$, and ${\Theta}^*_{unsup}$ are only theoretical values. On the finite data set $D$, $\hat{\Theta}_{smsup}^{S_l, S_u}$, $\hat{\Theta}_{usmsup}^{S_l,S_u}$ are the solutions to Eqs. \ref{eq:e4} and \ref{eq:e7}  respectively.
Let $\mathfrak{L}_f(\Theta)$ denote the loss of the Bayes plug-in classifier.

\begin{cor}
\label{cor2}
When $N_l \rightarrow \infty$, $N_u \rightarrow \infty$ and $N_l / N_u\rightarrow 0$, for the ideal solutions ${\Theta}^*_{sup}$,${\Theta}^*_{smsup}$,${\Theta}^*_{usmsup}$, and ${\Theta}^*_{unsup}$, there is,
\begin{equation}\small
\mathfrak{L}_f({\Theta}_{unsup}^*) = \mathfrak{L}_f({\Theta}_{smsup}^*) \geq \mathfrak{L}_f({\Theta}_{usmsup}^*) \geq \mathfrak{L}_f({\Theta}_{sup}^*)
\end{equation}
\end{cor}

%Thus, Eqs. \ref{eq:e4} and \ref{eq:e7} converge to different parameter set, i.e., supervised model parameters and unsupervised model parameters.
Thus, if $f$ is incorrect, the difference between $\hat{\Theta}_{usmsup}$ and $\hat{\Theta}_{smsup}$ will become larger and larger when more unlabeled data are used for model learning.

%Thus, Eqs. \ref{eq:e4} and \ref{eq:e7} converge to different parameter set, i.e., supervised model parameters and unsupervised model parameters. If $f$ is incorrect, then the difference between $\hat{\Theta}_{usmsup}$ and $\hat{\Theta}_{smsup}$ will become larger and larger when more training data are used. Let $\mathfrak{L}_f(\Theta)$ denote the loss of the Bayes plug-in classifier, then we have

\newtheorem{defn}{Definition}
\begin{defn}
With Corollay 2, if $\mathfrak{L}_f({\Theta}_{unsup}^*) > \mathfrak{L}_f ({\Theta}_{sup}^*)$, then $f$ is misspecified.
\end{defn}

\begin{thm}
if $\mathfrak{L}_f(\Theta^*_{unsup})>\mathfrak{L}_f(\Theta^*_{sup})$, then $\exists $  $S_l$, s.t.
\begin{equation}\small
\lim_{|S_u|\rightarrow \infty}P(\mathfrak{L}_f(\hat{\Theta}_{smsup}^{S_l,S_u})>\mathfrak{L}_f(\hat{\Theta}_{sup}^{S_l}))>0
\end{equation}
\end{thm}

This theorem shows that the semi-supervised learning yields degradation with a positive probability when more unlabeled data are introduced. The proof is straightforward and is not given due to the page limitation. %For the detail of proof, please referee \cite{Priebe:TPAMI:2011}.
By optimizing Eqs \ref{eq:regsemi} and \ref{eq:e7}, two distributions $f(X,Y|{\hat{\Theta}}_{smsup})$ and $f(X,Y|{\hat{\Theta}}_{usmsup})$ can be acquired. Then, the difference between the original and the unbiased semi-supervised learning can be defined as
\begin{equation}\small
\label{eq:diff}
KL(f(X,Y|{\hat{\Theta}}_{smsup})||f(X,Y|{\hat{\Theta}}_{usmsup}))
\end{equation}

\begin{thm}
\label{thm:gap}
If the model $f$ with  $\Theta$ is not misspecified,
%there is
%$\mathfrak{L}_f(\Theta^*_{sup})=\mathfrak{L}_f(\Theta^*_{unsup})$, \textbf{\emph{i.i.f.}}
\begin{equation}\small
\lim_{|S_u|\rightarrow \infty}
KL(f(X,Y|{\hat{\Theta}}_{smsup})||f(X,Y|{\hat{\Theta}}_{usmsup})) = 0
\end{equation}

\end{thm}

\textbf{Proof.} By \textbf{Definition} 1, if $f$ is not misspecified, there is $\mathfrak{L}_f({\Theta}_{unsup}^*) = \mathfrak{L}_f ({\Theta}_{sup}^*)$. Considering $\mathfrak{L}_f({\Theta}_{unsup}^*) \geq \mathfrak{L}_f({\Theta}_{smsup}^*)\geq \mathfrak{L}_f ({\Theta}_{sup}^*)$ and $\mathfrak{L}_f({\Theta}_{unsup}^*) \geq \mathfrak{L}_f({\Theta}_{usmsup}^*)\geq \mathfrak{L}_f ({\Theta}_{sup}^*)$, there is,
\begin{equation}\small
{\Theta}_{smsup}^* = {\Theta}_{usmsup}^* = {\Theta}_{sup}^* ={\Theta}_{unsup}^*
\end{equation}
So,
\begin{equation}\small
\begin{aligned}
\lim_{|S_u|\rightarrow \infty}
&KL(f(X,Y|{\hat{\Theta}}_{smsup})||f(X,Y|{\hat{\Theta}}_{usmsup})) \\ =& KL(f(X,Y|{{\Theta}}^*_{smsup})||f(X,Y|{{\Theta}^*}_{usmsup}))\\
=& \quad 0
\end{aligned}
\end{equation}

%The object of model modification is,\begin{equation}\label{eq:e14}\min_{\hat{\Theta}_{smsup},\hat{\Theta}_{usmsup}}KL(f(X,Y|{\hat{\Theta}}_{smsup})||f(X,Y|{\hat{\Theta}}_{usmsup}))\end{equation}

With the proposed theorems and corollaries, we have theoretically proved that the correctness of the previous conjectures illustrated in Figure \ref{fig:model} and a well-specified semi-supervised model-based classifier is proposed in the next Section.
%From Theorem \ref{thm:gap}, we can determine whether a model is misspecified or not by comparing $\Theta_{usmsup}$ with $\Theta_{smsup}$.
%%Intuitively, we can compute $KL(P_{k,\Theta_{ws}}||P_{k,\Theta_s})$ as the measurement. Provided with finite number of data samples, we can calculate the difference between $\Theta_{ub}^*$ and $\Theta_u^*$, $KL(P_{k,\Theta_{ws}}||P_{k,\Theta_s})$. If it is larger than a threshold, the the model is determined as misspecified.
%Combining these two semi-supervised strategies (Eqs \ref{eq:e4} and \ref{eq:e7}), the overall objective function is given as:
%
%\begin{equation}\small
%\label{eq:objective}
%\min_{\substack{\Theta_{usmsup}\\ \Theta_{smsup}}} \left\{
%             \begin{array}{lll}
%KL(P(X,{Y}||f(X,Y|\Theta_{smsup})) &+  \\
%KL(P(X,Y)||f(X,Y|\Theta_{usmsup}))&+\\
%KL(f(X,Y|\Theta_{smsup})||f(X,Y|\Theta_{usmsup}&))
%             \end{array}
%        \right\}
%\end{equation}

%\section{Adaptive Model Modification Based SSL}\label{sec:Approach}
\section{The Proposed ASKKM Approach}\label{sec:Approach}
To alleviate the model misspecification problem, we propose to adapt the model structure. In particular, we focus on kernel $k$-means model which can be considered as a special case of Gaussian mixture models (GMM), and explore mechanisms to learn the model structure and the model parameters to ensure the model to be well-specified.
%The model misspecification problem has three perspectives: (1) incorrect model assumption, i.e., the assumed model is inappropriate for the given data set, (2) inappropriate training data, and (3) incorrect model parameter assumption, e.g., the number of components for gaussian mixture model. Our approach is proposed for the third perspective.
Although different model complexity measures (such as BIC \cite{watanabe2013widely}) have been proposed to determine the optimal generative models like GMM, these measures were designed primarily for density estimation, and thus cannot be directly applied in the semi-supervised setting to address the misspecification problem. As discussed in Section \ref{sec:Problem}, we adopt the KL divergence between the original and unbiased semi-supervised learning to guide the adaptive model structure learning of a kernel $k$-means model. This section presents the proposed adaptive model modification based semi-supervised kernel $k$-means model (ASKKM for short).

\subsection{Model Misspecification Criterion}\label{sec:mm}

As illustrated in Figure \ref{fig:model}, if the assumed model is misspecified, the original SEM and unbiased SEM converge to different classification loss bound. The difference between classification loss bounds is approximated by the KL divergence between two SEMs. Practically, the discrete KL divergence might be
problematic when calculated on the limited number od data points. Therefore, the aggregated classification disagreement is adopted to approximate the bound difference. If the aggregated classification disagreement is large enough, then their KL divergence must be greater than 0 and thus exists model misspecification according to Theorem \ref{thm:gap}. Denote $\mathfrak{B}_{\hat{\Theta}_{usmsup}}(x)$ and $\mathfrak{B}_{\hat{\Theta}_{smsup}}(x)$ as the Bayes plug-in classifiers for original SEM and unbiased SEM, respectively. The criterion for model misspecification is defined as
%Operating Bayes plug-in classifiers for original SEM and unbiased SEM (denoted as $\mathfrak{B}_{\hat{\Theta}_{usmsup}}(x)$ and $\mathfrak{B}_{\hat{\Theta}_{smsup}}(x)$) on labeled training set $S_l$, two set of classification results are get. If there exists inconsistency between two classification results, we consider the difference between original SEM and unbiased SEM is too large, thus the model is misspecified following Theorem \ref{thm:gap}. So, by observing classification on $S_l$  the criterion of misspecification is given as,
%----------
\begin{equation}\label{eq:14}
Criterion = \sum_{x_i \in S_l} I(\mathfrak{B}_{\hat{\Theta}_{usmsup}}(x_i) \neq \mathfrak{B}_{\hat{\Theta}_{smsup}}(x_i))
\end{equation}
where $I(\cdot)$ is the indicator function. If $Criterion$ is greater than a predefined threshold $\epsilon$, the corresponding assumed model is determined as misspecified.

%With parameters $\hat{\Theta}_{usmsup}$ and $\hat{\Theta}_{smsup}$, the Bayes plug-in classifiers, denoted as $\mathfrak{B}_{\hat{\Theta}_{usmsup}}(x)$ and $\mathfrak{B}_{\hat{\Theta}_{smsup}}(x)$, can be acquired. %When evaluated on the labeled data set $S_l$, the classification results of unbiased  and original SSL can be acquired. %and their vector form are given as $\{\mathfrak{B}_{\hat{\Theta}_{usmsup}}(x_1),\cdots, \mathfrak{B}_{\hat{\Theta}_{usmsup}}(x_{N_l})\}$ and $\{\mathfrak{B}_{\hat{\Theta}_{smsup}}(x_1),\cdots, \mathfrak{B}_{\hat{\Theta}_{smsup}}(x_{N_l})\}$.
%If there exist $x_i \in S_l$, such that $\mathfrak{B}_{\hat{\Theta}_{usmsup}}(x_i) \neq \mathfrak{B}_{\hat{\Theta}_{smsup}}(x_i)$, we consider the difference between $\hat{\Theta}_{usmsup}$ and $\hat{\Theta}_{smsup}$ is too large, thus the model is misspecified following Theorem \ref{thm:gap}.

If model misspecification occurs, we gradually increase the model complexity of the employed semi-supervised generative model by modifying K, i.e., the number of components. Specially, for each labeled training data $x_i \in S_l$, a new label $c_i$ is assigned to it if $Criterion > \epsilon$. The size of new label set $C$ is then larger than that of the given class label set $Y$, i.e., $|C| > |Y|$. For the classification task, a mapping function from new label set to the given label set is defined as,
\begin{equation}   g(c_i) =
 \begin{cases}
    y_i   &  \text{if} \quad \mathfrak{B}_{\hat{\Theta}_{usmsup}}(x_i) = \mathfrak{B}_{\hat{\Theta}_{smsup}(x_i)} \\
   \mathfrak{B}_{\hat{\Theta}_{usmsup}}(x_i)
        &  \text{if} \quad \mathfrak{B}_{\hat{\Theta}_{usmsup}}(x_i) \neq \mathfrak{B}_{\hat{\Theta}_{smsup}(x_i)}
 \end{cases}                \end{equation}

With this function, a new cluster is introduced for the new class label and thus the model structure is adaptively modified. %number of clusters is dynamically increased. %After the number (K) of clusters is changed, the corresponding model parameters for each cluster should be relearnt during the model learning process.

\renewcommand{\algorithmicrequire}{\textbf{Input:}}

\renewcommand{\algorithmicensure}{\textbf{Output:}}

\begin{algorithm} \small   \caption{Adaptive Semi Kernel $k$-means Model}    \label{algorithm:hkkm}
\begin{algorithmic}[1]
\STATE initialize $K$ with number of the natural classes, $K=|Y|$;
\STATE initialize the assignment of unlabeled samples $S_u$ based on the kernel distance;
\WHILE{no convergence}
\STATE update Kernel maps and the centroids for original SSL and unbiased SSL respectively;
\STATE update cluster assignments $Z^1$,  $Z^2$;
\IF{$Z^1_l \neq Z^2_l$}
\STATE modify the misspecified model and update $K$;
\STATE  \textbf{goto} 2;
\ENDIF
\ENDWHILE
\end{algorithmic}
\end{algorithm}

\begin{figure}[!htb]
\centering
\includegraphics[width=3.0in]{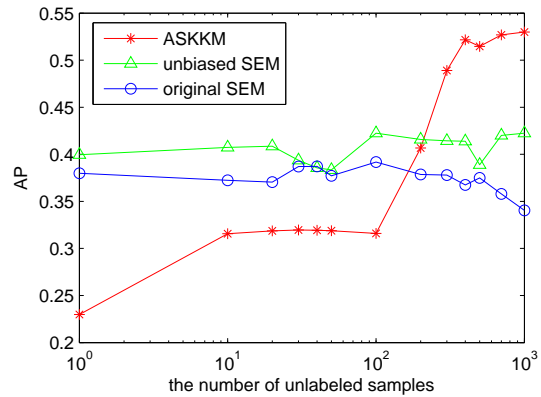}
\caption{The effect of modifying model structures. }
\label{fig:abVSl}
\end{figure}

%
%\begin{figure}[!htb]
%\centering
%\includegraphics[width=3.3in]{apVSk.eps}
%\caption{The effect on modifying K.}
%\label{fig:apVSk}
%\end{figure}
%%\vspace{-0.2cm}

\begin{table*}[htbp]\small
  \centering
  \caption{Results on PASCAL Dataset}
    \begin{tabular}{l|ccccc|cc}
    \toprule
    \multirow{2}[4]{*}{\textbf{PASCAL07}} & \multicolumn{5}{c|}{\textbf{without tag}} & \multicolumn{2}{c}{\textbf{with tag}} \\
\cmidrule{2-8}          & \textbf{S4VM} & \textbf{co-training} & \textbf{original SEM} & \multicolumn{1}{r}{\textbf{unbiased SEM}} & \textbf{ASKKM} & \textbf{MKL+tag} & \textbf{ASKKM+tag} \\
    \midrule
    \textbf{aeroplane} & 0.443 & 0.467 & 0.413 & 0.355 & \textbf{0.468} & \textbf{0.592} & 0.524 \\
    \textbf{bicycle} & 0.182 & 0.233 & 0.146 & 0.148 & \textbf{0.331} & 0.324 & \textbf{0.361} \\
    \textbf{bird} & \textbf{0.229} & 0.221 & 0.191 & 0.22  & 0.196 & 0.376 & \textbf{0.451} \\
    \textbf{boat} & 0.301 & 0.315 & 0.227 & 0.259 & \textbf{0.409} & \textbf{0.519} & 0.478 \\
    \textbf{bottle} & \textbf{0.211} & 0.203 & 0.166 & 0.166 & 0.204 & 0.154 &\textbf{ 0.177} \\
    \textbf{bus} & 0.294 & 0.253 & 0.159 & 0.277 & \textbf{0.339} & 0.278 & \textbf{0.481} \\
    \textbf{car} & 0.441 & 0.457 & 0.353 & 0.438 & \textbf{0.616} & 0.501 & \textbf{0.533} \\
    \textbf{cat} & 0.281 & 0.265 & 0.203 & 0.259 & \textbf{0.321} & 0.366 & \textbf{0.402} \\
    \textbf{chair} & 0.371 & 0.318 & 0.322 & 0.332 & \textbf{0.394} & 0.3   & \textbf{0.388} \\
    \textbf{cow} & 0.162 & 0.175 & 0.135 & 0.159 & \textbf{0.199} & 0.117 & \textbf{0.159} \\
    \textbf{diningtable} & 0.242 & 0.232 & 0.167 & 0.221 & \textbf{0.277} & 0.255 & \textbf{0.289} \\
    \textbf{dog} & 0.308 & 0.279 & 0.213 & 0.313 & \textbf{0.316} & 0.331 & \textbf{0.401} \\
    \textbf{horse} & 0.265 & 0.278 & 0.182 & 0.181 & \textbf{0.437} & 0.637 & \textbf{0.697} \\
    \textbf{motorbike} & 0.257 & 0.228 & 0.148 & 0.251 & \textbf{0.361} & 0.383 & \textbf{0.428} \\
    \textbf{person} & 0.654 & 0.638 & 0.551 & 0.596 & \textbf{0.696} & \textbf{0.703} & 0.617 \\
    \textbf{pottedplant} & 0.174 & 0.175 & 0.162 & 0.183 & \textbf{0.191} & \textbf{0.212} & 0.186 \\
    \textbf{sheep} & 0.208 & \textbf{0.212} & 0.145 & 0.163 & 0.169 & 0.218 & \textbf{0.297} \\
    \textbf{sofa} & 0.218 & 0.251 & 0.171 & 0.181 & \textbf{0.266} & 0.191 & \textbf{0.201} \\
    \textbf{train} & 0.416 & 0.394 & 0.182 & 0.309 & \textbf{0.421} & 0.617 & \textbf{0.683} \\
    \textbf{TVmonitor} & 0.261 & 0.228 & 0.172 & 0.223 & \textbf{0.331} & 0.236 & \textbf{0.344} \\
    \hline
    \textbf{mAP} & 0.296 & 0.291 & 0.221 & 0.262 & \textbf{0.347} & 0.366 & \textbf{0.405} \\
    \hline
    \end{tabular}%
  \label{tab:resultl}%
\end{table*}%

\begin{table*}[!ht]\small
  \centering
  \caption{Results on MIR Flickr Dataset}
    \begin{tabular}{l|ccccc|cc}
    \toprule
    \multirow{2}[4]{*}{\textbf{MIR Flickr}} & \multicolumn{5}{c|}{without tag}      & \multicolumn{2}{c}{with tag} \\
\cmidrule{2-8}          & \textbf{S4VM} & \textbf{co-training} & \textbf{original SEM} & \textbf{unbiased SEM} & \textbf{ASKKM} & \textbf{MKL+tag} & \textbf{ASKKM+tag} \\
    \midrule
    \textbf{animals} & 0.266 & \textbf{0.311} & 0.256 & 0.256 & 0.286 & \textbf{0.31} & 0.287 \\
    \textbf{baby} & \textbf{0.132} & 0.122 & 0.114 & 0.121 & 0.109 & 0.075 & \textbf{0.153} \\
    \textbf{baby*} & 0.154 & 0.128 & 0.121 & 0.122 & \textbf{0.163} & 0.161 & \textbf{0.184} \\
    \textbf{bird} & 0.131 & \textbf{0.136} & 0.125 & 0.131 & 0.126 & 0.124 & \textbf{0.139} \\
    \textbf{bird*} & 0.127 & 0.131 & 0.121 & 0.119 & \textbf{0.143} & 0.163 & \textbf{0.217} \\
    \textbf{car} & 0.221 & 0.184 & 0.141 & \textbf{0.227} & \textbf{0.227} & \textbf{0.229} & 0.223 \\
    \textbf{car*} & 0.156 & 0.212 & 0.138 & 0.174 & \textbf{0.249} & \textbf{0.305} & 0.263 \\
    \textbf{clouds} & 0.585 & 0.641 & 0.421 & 0.501 & \textbf{0.676} & 0.612 & \textbf{0.682} \\
    \textbf{clouds*} & 0.459 & 0.53  & 0.271 & 0.482 & \textbf{0.613} & 0.537 & \textbf{0.677} \\
    \textbf{dog} & 0.165 & 0.157 & 0.133 & 0.146 & \textbf{0.166} & 0.182 & \textbf{0.276} \\
    \textbf{dog*} & 0.183 & \textbf{0.202} & 0.134 & 0.133 & 0.138 & 0.212 & \textbf{0.315} \\
    \textbf{female} & 0.443 & 0.438 & 0.405 & \textbf{0.466} & 0.391 & 0.44  & \textbf{0.441} \\
    \textbf{female*} & 0.364 & \textbf{0.377} & 0.293 & \textbf{0.377} & 0.375 & 0.313 & \textbf{0.389} \\
    \textbf{flower} & 0.258 & 0.279 & 0.253 & 0.233 & \textbf{0.336} & 0.373 & \textbf{0.413} \\
    \textbf{flower*} & 0.276 & 0.271 & 0.223 & 0.266 & \textbf{0.3} & \textbf{0.424} & 0.391 \\
    \textbf{food} & 0.248 & 0.26  & 0.139 & 0.179 & \textbf{0.321} & 0.333 & \textbf{0.354} \\
    \textbf{indoor} & 0.529 & \textbf{0.554} & 0.515 & 0.525 & 0.523 & 0.514 & \textbf{0.542} \\
    \textbf{lake} & 0.207 & 0.214 & 0.218 & 0.211 & \textbf{0.223} & 0.159 & \textbf{0.244} \\
    \textbf{male} & 0.401 & 0.417 & 0.315 & \textbf{0.423} & 0.346 & 0.366 & \textbf{0.385} \\
    \textbf{male*} & \textbf{0.327} & 0.303 & 0.287 & 0.308 & 0.321 & 0.255 & \textbf{0.283} \\
    \textbf{night} & \textbf{0.429} & 0.437 & 0.189 & 0.326 & 0.383 & \textbf{0.471} & 0.436 \\
    \textbf{night*} & 0.304 & 0.301 & 0.135 & 0.211 & \textbf{0.321} & 0.368 & \textbf{0.426} \\
    \textbf{people} & \textbf{0.649} & 0.629 & 0.604 & 0.582 & 0.604 & 0.629 & \textbf{0.671} \\
    \textbf{people*} & 0.554 & \textbf{0.562} & 0.516 & 0.532 & 0.535 & 0.554 & \textbf{0.597} \\
    \textbf{plant life} & 0.547 & 0.63  & 0.491 & 0.486 & \textbf{0.636} & 0.613 & \textbf{0.643} \\
    \textbf{portrait} & 0.414 & 0.443 & 0.389 & 0.421 & \textbf{0.454} & \textbf{0.474} & 0.441 \\
    \textbf{portrait*} & \textbf{0.448} & 0.406 & 0.402 & 0.334 & 0.437 & \textbf{0.429} & 0.423 \\
    \textbf{river} & 0.194 & 0.205 & 0.184 & 0.195 & \textbf{0.218} & 0.234 & \textbf{0.295} \\
    \textbf{river*} & 0.117 & 0.118 & 0.118 & \textbf{0.124} & 0.051 & 0.047 & \textbf{0.094} \\
    \textbf{sea} & 0.374 & 0.321 & 0.334 & 0.408 & \textbf{0.448} & \textbf{0.437} & \textbf{0.437} \\
    \textbf{sea*} & 0.184 & \textbf{0.193} & 0.177 & 0.135 & 0.177 & 0.255 & \textbf{0.302} \\
    \textbf{sky} & 0.642 & 0.647 & 0.603 & 0.589 & \textbf{0.719} & 0.693 & \textbf{0.726} \\
    \textbf{structures} & 0.658 & 0.652 & 0.602 & 0.651 & \textbf{0.659} & 0.655 & \textbf{0.693} \\
    \textbf{sunset} & 0.374 & 0.368 & 0.226 & 0.342 & \textbf{0.416} & \textbf{0.543} & 0.487 \\
    \textbf{transport} & 0.309 & 0.295 & 0.286 & 0.289 & \textbf{0.326} & 0.321 & \textbf{0.395} \\
    \textbf{tree} & 0.437 & \textbf{0.485} & 0.375 & 0.418 & 0.469 & 0.453 & \textbf{0.461} \\
    \textbf{tree*} & 0.209 & \textbf{0.265} & 0.175 & 0.228 & 0.234 & 0.231 & \textbf{0.326} \\
    \textbf{water} & 0.428 & 0.437 & 0.396 & 0.451 & \textbf{0.495} & 0.452 & 0.513 \\
    \midrule
    \textbf{mAP} & 0.339 & 0.348 & 0.285 & 0.319 & \textbf{0.359} & 0.367 & \textbf{0.401} \\
    \bottomrule

    \end{tabular}%
  \label{tab:result2}%
\end{table*}%

\subsection{Adaptive Semi-supervised Kernel K-means Model} \label{sec:askk}
%The aim of this paper is to modify misspecified models and train an adaptive SSL model. Combining the original SEM (Eq.\ref{eq:e4}), the unbiased SEM (Eq.\ref{eq:e7}) with the objective function of model modification (Eq.\ref{eq:e14}),  the overall objective function of our model is given as,
%\begin{equation}\small
%\label{eq:objective}
%\min_{\substack{\Theta_{usmsup}\\ \Theta_{smsup}}} \left\{
%             \begin{array}{lll}
%KL(P(X,{Y}||f(X,Y|\Theta_{smsup})) &+  \\
%KL(P(X,Y)||f(X,Y|\Theta_{usmsup}))&+\\
%KL(f(X,Y|\Theta_{smsup})||f(X,Y|\Theta_{usmsup}&))
%             \end{array}
%        \right\}
%\end{equation}
For the classification task, a kernel $k$-means is adopted in the paper, and accordingly Eqs. \ref{eq:e4} (original SEM) and Eq.\ref{eq:e7} (unbiased SEM) can be rewritten respectively as
\begin{equation}
\label{eq:osk}\small
\begin{aligned}
\{\hat{\varphi}_1, \hat{Z}^1 \}=&\argmin_{\varphi_1,Z^1}  \{\sum_{x_i \in S_l} \parallel \varphi_1(x_i) -\varphi_1(\mu_1^k) \parallel^2 +\\ & \sum_{x_i \in S_u} z^1_{i,k}\parallel\varphi_1(x_i) - \varphi_1(\mu_1^k)\parallel^2\}
\end{aligned}
\end{equation}

\begin{equation}
\label{eq:wsk}\small
\begin{aligned}
\{\hat{\varphi}_2, \hat{Z}^2 \}=&\argmin_{\varphi_2,Z^2}  \{\sum_{x_i \in S_l} \parallel \varphi_2(x_i) -\varphi_2(\mu_2^k) \parallel^2 +\\ & \frac{N_l}{N_u+N_l}\sum_{x_i \in S_u} z^2_{i,k}\parallel\varphi_2(x_i) - \varphi_2(\mu_2^k)\parallel^2\}
\end{aligned}
\end{equation}
%%where $\varphi_s,\mu_s^k$, are the kernel maps and centroids respectively, $s=1,2$, and $Z^s$  are the cluster assignments with $Z^s = Z^s_l \cup Z^s_u$. That is, if $x_i$ is assigned to the $k$-th cluster, $z^s_{i,k} = 1$.

where $\varphi_1,\mu_1^k$ and $\varphi_2,\mu_2^k$ are respectively the kernel maps and the centroids for the original and weighted semi-supervised kernel $k$-means, and $Z^1$ and $Z^2$ are the cluster assignments with $Z^1 = Z^1_l \cup Z^1_u$ and $Z^2 = Z^2_l \cup Z^2_u$. That is, if $x_i$ is assigned to the $k$-th cluster according to Eq. \ref{eq:osk} or \ref{eq:wsk}, $z^1_{i,k} = 1$ or $z^2_{i,k} = 1$. Intuitively speaking, the proposed approach tracks the difference between $Z_l^1$ and $Z_u^2$. If $Z_l^1 \neq Z_u^2$, Eq. \ref{eq:14} is used to check whether model misspecification occurs or not. Details of the proposed ASKKM is illustrated in Algorithm \ref{algorithm:hkkm}.

%Following the objective function as Eq.\ref{eq:objective}, the first two terms are minimized by iteration of $k$-means and the third term is minimized by model modification. That is,
%If $Z_l^1 \neq Z_l^2$, the criterion as Eq. \ref{eq:14} is used to check model misspecification. After model modification, reset parameter and run $k$-means. Details of the proposed ASKKM is illustrated in Algorithm \ref{algorithm:hkkm}.

%\begin{algorithm}
%    \caption{Adaptive Semi Kernel $k$-means Model}
%    \label{algorithm:hkkm}
%    \begin{algorithmic}[1]
%
%\STATE Initialize $K=C$, and $K,C$ is the number of clusters and classes, respectively;
%\STATE Using $S_l$ to initialize $S_u$ based on the kernel distance;
% \WHILE{not converge}
% \STATE update the kernel map $\varphi_1$, $\varphi_2$ and the centroids;
% \STATE update the cluster assignments $Z^1$,  $Z^2$;
% \IF{$Z^1_l \neq Z^2_l$}
% \STATE modify the model by adding new class $c_i$ and update $K=K+1$;
% \STATE  \textbf{goto} 2;
% \ENDIF
%\ENDWHILE
%\end{algorithmic}
%\end{algorithm}

\section{Experimental Results}\label{sec:Exp}

For experimental evaluation, we evaluate the proposed ASKKM using two image classification data sets, i.e., PASCAL VOC'07 \cite{Everingham2010The} and MIR Flickr \cite{Huiskes2008The}. PASCAL VOC'07 consists of 9,963 images from 20 classes and 804 annotated tags. Among them, 5,011 images are selected as the training set and the rest form the test set. MIR Flikr contains 25,000 images and 457 tags from 38 classes collected from the Flickr website. Among them, 12,500 images are randomly selected for training and the rest form the test set. For image feature representations, two local features (SIFT, Hue), three global histogram features (RGB, Hsv and Lab) as well as GIST are used to represent each image. We adopted different distance metrics for features of different types. In particular, the Manhattan distance, Euclidean distance and Chi-square distance are used respectively for the histogram features, the GIST features and the local features. The state-of-the-art SSL algorithms as well as semi-supervised generative models are chosen for model comparison, including S4VM \cite{Li2011Towards}, co-training \cite{Zhou2005Semi}, semi-supervised EM (SEM) \cite{Fox-Roberts:2014:JMLR} and MKL \cite{Guillaumin2010Multimodal}. In addition to utilizing the labels of images, the original MKL also utilizes the tags of images. Therefore, we extend the proposed ASKKM in the same way as what MKL does to utilize the tag information.
% and then only compared with MKL with tags.
For performance evaluation metric, we adopt average precision (AP) which is the evaluation criterion used in PASCAL VOC competition, written as
%-----------------
\begin{eqnarray}\small
\begin{aligned}
\nonumber & AP = \frac{1}{11}\sum_{r \in {0,0.1,...,1}}{P_{interp}(r)}\\
\nonumber & P_{interp}(r) = \max_{\tilde{r}:\tilde{r}\geq r} p{\tilde{r}}
\end{aligned}
\end{eqnarray}
%-----------------
where this criterion requires the recall $r$ to take value from 0 to 1 with the step as 0.1, and then sums up the precision over all $r$ and takes the average value. The performance comparison results are reported in the following sections.

%\subsection{The effect on the number of clusters}
%
%In this section, we first verify the effect of the proposed model modification strategy. The verification is performed on the sub data set of ``clouds'' class in MIR data set, and the number of labeled training data is 50. In the experiments, $K$ is fixed to 2 for all approaches at the beginning. Then, the proposed ASKKM keeps on adapting the model structure (K) once the model modification occurs, while the original SEM and unbiased SEM keep unchanged.
%
%Experimental results are plotted in Figure \ref{fig:apVSk}. From this figure, it is noticed that at the beginning stage of the curve when model misspecification occurs, the $AP$ rapidly increase from 0.3 to 0.65. When $K \geq 6$, AP value gradually converges and achieves the best AP (0.7) when $K=10$. While the unbiased SEM is better than the original SEM which is 0.5 vs 0.45, and both are worse than the ASKKM. This result verifies the effectiveness of varying $K$, i.e., the adaption of model structure.

\subsection{Verification of Model Structure Modification}

To evaluate how the adaptive modification on model structure could affect the model performance, experiments are performed on ``car'' data set of PASCAL VOC'07. For this binary classification task, $K$ is fixed to 2 for the original SEM and ubiased SEM model \cite{Fox-Roberts:2014:JMLR}. However, the true data distributions for image data may contain more than two clusters (components). Therefore, the proposed ASKKM adaptively modifies $K$ once the model misspecification is detected during the experiments. The comparison results are then plotted in Figure \ref{fig:abVSl}.

From this figure, it is noticed that the model performance of the original SEM gradually degrades after $N_u > 100$. The unbiased SEM is better than the original SEM as its AP value, although slightly fluctuates, almost keeps around 0.4 which does not degrade. For the performance of the ASKKM, it slowly increases at the beginning part of the curve. After $N_u>100$, its AP value dramatically increases where model misspecification is detected and the model structure is accordingly modified. Then, the ASKKM gradually converge with the addition of unlabeled data. The converged model performance of ASKKM is much better than that of the compared semi-supervised generative models. This verifies that a well-specified semi-supervised models could acquire a superior model performance.

\subsection{Model Performance Comparison}

The model performance evaluation is performed on aforementioned two image data sets. For each class in the data set, we only choose few labeled data, i.e., $N_l=20$ for PASCAL VOC'07 data set and $N_l=50$ for MIR Flikr data set. We not only compare our approach with two semi-supervised generative models but also with the most representative semi-supervised learning algorithms such as S4VM and co-training. Experimental results are reported in Table \ref{tab:resultl} and Table \ref{tab:result2}.

From Table \ref{tab:resultl}, the model performance of the ASKKM is the best on 17 classes out of 20 classes when tag information is not considered. The S4VM achieves the best AP value in class ``aeroplane'' and ``bottle''. The MAP value of the ASKKM is $17.2\%$ higher than the second best model S4VM. It is also noticed that semi-supervised SVM based algorithms performs better than the generative model based ones. The superior performance of the proposed approach indicates the superiority of the ASKKM over the rest approaches. If tag information is considered, it is observed that MKL+tag is better than these approaches without the integration of tag information, and this is consistent with our intuition. However, the ASKKM+tag is better than MKL+tag in 16 classes and the overall MAP of the ASKKM is $10.6\%$ higher than that of MKL+tag. This further verify the effectiveness of the proposed approach. Similar observations could be found in the evaluation results on MIR Flickr data set reported in Table \ref{tab:result2}. From these rigorous experimental results, we can conclude the proposed ASKKM is superior to the state-of-the-art semi-supervised learning approaches in terms of average precision and mean average precision.

\section{Conclusion}\label{sec:Conclusion}

To learn a reliable semi-supervised models is of utmost importance. Most of existing works are non-parametric based ones and the generative model based approach is seldom studied. This paper first proposes a criterion to judge whether a model misspecificatoin occurs or not. Then an adaptive semi-supervised kernel K-means model (ASKKM) is proposed for the model misspecified problem. At last, we rigorously evaluate the proposed ASKKM on two image classification data sets, i.e., PASCAL VOC'07 and MIR Flickr. Promising results demonstrate the efficacy of the proposed approach.

%%% The file named.bst is a bibliography style file for BibTeX 0.99c
\bibliographystyle{named}
\bibliography{ijcai17}

\end{document}